\documentclass[10pt,twocolumn,letterpaper]{article}

\usepackage{3dv}

\usepackage{times}
\usepackage{epsfig}
\usepackage{graphicx}
\usepackage{amsmath}
\usepackage{amssymb}
\usepackage{booktabs}
\usepackage{pifont}
\usepackage{comment}
\usepackage{multirow}
\usepackage{microtype}
\usepackage{xspace}
\usepackage{stfloats}
\usepackage[pagebackref=true,breaklinks=true,colorlinks,bookmarks=false]{hyperref}

\usepackage{soul}
\usepackage{amsmath}
\usepackage{amssymb}
\usepackage{microtype}
\usepackage{wrapfig}
\usepackage{fancyhdr}
\usepackage{xcolor}

\def\figurePath{figures/}
\def\myfigure#1#2{\begin{figure}[htb]\centering\includegraphics*[width = \linewidth]{\figurePath#1}\centering\vspace{-0.0cm}\caption{#2}\label{fig:#1}\end{figure}}

\def\mycfigure#1#2{\begin{figure*}[t]\centering\includegraphics*[clip, width = \linewidth]{\figurePath#1}\centering\vspace{-0.0cm}\caption{#2}\label{fig:#1}\end{figure*}}

\newcommand{\argmin}[1]{\underset{#1}{\operatorname{arg\,min}\ }}

\newcommand{\refSec}[1]{Sec.~\ref{sec:#1}}
\newcommand{\refFig}[1]{Fig.~\ref{fig:#1}}
\newcommand{\refEq}[1]{Eq.~\ref{eq:#1}}
\newcommand{\refTbl}[1]{Tbl.~\ref{tbl:#1}}

\soulregister\ref7
\soulregister\cite7
\soulregister\refFig7
\soulregister\cite7
\soulregister\ref7
\soulregister\pageref7
\soulregister\shortcite7
\soulregister\eg0
\soulregister\ie0
\soulregister\etal0

\DeclareGraphicsExtensions{.png,.jpg,.pdf,.ai,.psd}
\newcommand{\mysection}[2]{\section{#1}\label{sec:#2}}
\newcommand{\mysubsection}[2]{\subsection{#1}\label{sec:#2}}

\newcommand{\iconA}{\scalebox{0.8}{\ding{108}}}
\newcommand{\iconB}{\scalebox{0.8}{\ding{117}}}
\newcommand{\iconC}{\scalebox{0.8}{\ding{116}}}

\newcommand{\iconF}{\ding{72}}
\newcommand{\iconG}{\ding{89}}
\newcommand{\iconH}{\ding{88}}

\definecolor{colorA}{HTML}{4285f4}
\definecolor{colorB}{HTML}{ea4335}
\definecolor{colorC}{HTML}{fbbc04}
\definecolor{colorD}{HTML}{34a853}
\definecolor{colorE}{HTML}{ff6d01}
\definecolor{colorF}{HTML}{46bdc6}
\definecolor{colorG}{HTML}{000000}
\definecolor{colorH}{HTML}{777777}

\newcommand{\colorIcon}[2]{\textcolor{color#1}{\csname icon#2\endcsname}}
\newcommand{\colorIconAndName}[2]{\colorIcon{#1}{#2}~\text{#2}}
\newcommand{\nameInColor}[1]{\textcolor{color#1}{\texttt{#1}}}
\newcommand{\shortNameInColor}[2]{\textcolor{color#1}{\texttt{#2}}}
\newcommand{\nameAndIcon}[1]{``#1'' (denoted~{\csname icon#1\endcsname})}

\newcommand{\myparagraph}[1]{\textbf{#1}}
\newcommand{\iconCircles}{\iconA}
\newcommand{\iconSpheres}{\iconB}
\newcommand{\iconChairs}{\iconG}
\newcommand{\iconMulti}{\iconC}
\newcommand{\iconVaried}{\iconH}
\newcommand{\iconLight}{\iconF}
\newcommand{\nacell}{\multicolumn1c{---}}
\newcommand{\dataset}[1]{\textsc{#1}}

\newcommand{\cmark}{\checkmark}%
\newcommand{\xmark}{\scalebox{0.85}{\ding{53}}}%
\newcommand{\mymath}[2]{\newcommand{#1}{\TextOrMath{$#2$\xspace}{#2}}}
\newcommand{\bestSupervised}[1]{\textbf{#1}}
\newcommand{\bestUnsupervised}[1]{\textbf{#1}}

\mymath{\network}{G}
\mymath{\image}{I}
\mymath{\sceneCode}{\mathbf s}
\mymath{\rendering}{\mathcal R}
\mymath{\encoder}{E}
\mymath{\latentCode}{\mathbf z}
\mymath{\critic}{C}
\mymath{\loss}{\mathcal L}
\mymath{\parameters}{\theta}
\mymath{\weight}{\lambda}
\mymath{\sceneCodeSize}{n}

\colorlet{colorSuper5}{colorA}
\colorlet{colorSuper10}{colorB}
\colorlet{colorNonCur}{colorC}
\colorlet{colorOur}{colorD}

\threedvfinalcopy
\def\threedvPaperID{232}

\begin{document}

\title{Curiosity-driven 3D Object Detection Without Labels}

\author{David Griffiths, Jan Boehm, Tobias Ritschel\\
University College London\\
{\tt\small \{david.griffiths.16, j.boehm, t.ritschel\}@ucl.ac.uk}
}

\maketitle
\thispagestyle{empty}

\begin{abstract}
In this paper we set out to solve the task of 6-DOF 3D object detection from 2D images, where the only supervision is a geometric representation of the objects we aim to find. In doing so, we remove the need for 6-DOF labels (\ie position, orientation etc.), allowing our network to be trained on unlabeled images in a self-supervised manner. We achieve this through a neural network which learns an explicit scene parameterization which is subsequently passed into a differentiable renderer. We analyze why analysis-by-synthesis-like losses for supervision of 3D scene structure using differentiable rendering is not practical, as it almost always gets stuck in local minima of visual ambiguities. This can be overcome by a novel form of training, where an additional network is employed to steer the optimization itself to explore the entire parameter space \ie to be curious, and hence, to resolve those ambiguities and find workable minima.\end{abstract}

\mysection{Introduction}{Introduction}
Deep learning has provided impressive abilities to infer 3D understanding from 2D images alone.
However, comes at the cost of manual labeling of 3D supervision, preventing application to the ``heavy tail'' of important tasks where no 3D scene supervision (position, orientation, color etc.) is available, and users have to make do with no more than a set of 2D images.

\myfigure{Teaser}{Trained on no more than a set of 2D images containing 3D objects with known geometry, without any labels, our approach learns a network to map 2D images to 3D scene parameters such as position, orientation, color, size, type, illumination, etc., which can be easily manipulated or used directly for downstream 3D scene understanding tasks.}

For this reason, there has recently been a push to get 3D with 2D-only self-supervision \cite{kato2018neural,chen2019learning,henzler2019platonicgan,tulsiani2017multi,henderson2019learning,han2020drwr}. In this paper, we address the common situation where the geometry of the object is known, however, scene labels are not available. Such scenarios are common in industrial environments where components are designed using Computer Aided Design (CAD) software, as well as scenarios where object geometry is consistent (\ie cars, street furniture etc.). Armed with just a 3D geometric representation and unlabeled 2D images, we aim to recover the position, orientation, color, and illumination for every relevant object in the scene.

Our approach uses a common Convolutional Neural Network (CNN) and Multi-Layer Perceptron (MLP) network to map an image to an explicit and interpretable 3D scene code that controls a Differentiable Renderer (DR).
This scene code, like a scene graph, has direct meaning and can be used in other downstream tasks such as loading it into a 3D graphics application for manipulation or re-synthesis with many applications in augmented and virtual reality.
Alternatively, the scene code itself can be used directly for 3D scene understanding tasks such a 3D object detection.

%\myparagraph{Problem statement}
This problem is difficult, as unfortunately, direct application of a DR for self-supervised learning will not result in a reliable optimization.
Consider (\refFig{Concept}, left), a typical 3D scene made of the 3D position of a quad and its RGB color, an apparently simple 6D problem, and further consider this is to be learned from 2D images of the 3D quad in front of a solid background.
Now, if gradient descent starts to minimize a $\mathcal L_2$-like error ($\mathcal L_1, \mathcal L_2$, Huber norm, etc.), in almost every case, the initial guess is far off the right values (``Iter 1'' to ``Iter $n$'' in \refFig{Concept}).
To satisfy a $\mathcal L_2$-like loss, the best thing to do is to make the quad small in 2D, \eg by moving it away from the viewpoint, and have the color become the one of the background at that image position, even if it is not at all the position of the quad in the image and neither its color.
This is just one of the many ambiguities that exist in the mapping between scene parameters and images.
Essentially, in almost every position of the optimization space, following the gradient leads to an unusable solution.
Furthermore, even if for one optimization step there is an overlap and the gradient was meaningful, it will be followed by too many bad ones to be useful.

We found this problem can be overcome by enforcing samples generated by the DR to match the distribution of the input dataset.
We implement this by introducing a second network, akin to a critic (or discriminator) in adversarial training \cite{goodfellow2014gan}. 
The by-the-book example for adversarial training is super-resolution \cite{ledig2017photo}, a low-resolution patch can be explained by many high-resolution patches, but their average, which a network without a critic will find, is not a valid high-resolution patch.
In our case, however, the degenerate solutions (\eg small ``mini chameleon'' quads hiding in the back) are valid samples from the data distribution.
Nothing is wrong with an individual sample.
They are also not an average of many solutions, they are not blurry, they are just small quads as they occur in the data distribution.
The key is, that the overall distribution of such solutions, rendered back to an image, is far from the original data distribution.
Hence, a critic will push the optimization to not rely on the same (and incorrect) answer all the time and be ``curious'' instead.
It forces the optimizer to continue looking for a better solution within the data distribution.
Eventually, the optimization discovers a solution that is not always the same, and ultimately, even more correct in the $\mathcal L_2$-like sense.
The critic avoids the ``easy reward'' from following the gradient of the $\mathcal L_2$-like loss.
Our instrumentation assures this cannot be achieved by just weighting $\mathcal L_2$ differently, by using higher learning rate, more randomization, longer learning, or other hyper-parameters.

\mycfigure{Concept}{Curiosity-driven and direct learning:
This simple world comprises of quads with different positions and depth as well as varying color.
Starting from a set of 2D images (three shown), we could train a mapping to directly produce 3D scene parameters (position and color of the quad).
Unfortunately, $\mathcal L_2$ rendering gradients will make the optimization converge (blue plot) to the wrong solution of very small quads with the color of the background in almost all cases.
Our curiosity-driven approach adds a critic to look at the re-rendering of the 3D scene parameters, forcing the optimization to keep trying unless it finds a solution that matches the 2D image data distribution (red plot).
Such learned parameters start to have meaningful gradients, and following them, ultimately, leads to a better solution (star).
}

Whilst our scene parameterization can be used for many downstream tasks, we evaluate our method on the most direct application of the parameterization itself, 3D object detection. Specifically, we tackle the problem of 3D object detection from monocular 2D images learned from 2D-only self-supervision with known geometry. We evaluate our approach on increasingly complex synthetic scenes as well as real scenes where the background is constant. Finally, we demonstrate the effectiveness of our approach where the background is also unknown by randomly generating training data from the full distribution of possible parameters using a simple rendering pipeline on cluttered backgrounds. We find such an approach achieves state-of-the-art results for unsupervised methods on the LineMOD dataset \cite{hinterstoisser2011multimodal}.

\mysection{Previous Work}{PreviousWork}

% Supervised
\myparagraph{Supervised}
Monocular object detection from 2D to 3D is notoriously hard due to being an ill-posed problem, owed to the projective entanglement of depth and scale. Despite this, many methods have been proposed to map a single image to set of 3D objects. Typical approaches rely on a dense set of labels for supervision \cite{liu2020smoke, wang2020centernet3d, zhou2021monocular}. \cite{chen2016monocular} show many different cues such as masks, absolute locations and shape as well as priors, such as assuming a known ground plane can be effective in 2D-3D. \cite{mousavian20173d} leverage 2D bounding box detection, by estimating 3D cuboids from the former. Lifting 2D to 3D has proven to be very successful \cite{Simonelli_2019_ICCV}. However, these methods are limited to cases where supervision labels are available (\eg for autonomous driving \cite{geiger2013vision, sun2020scalability, caesar2020nuscenes}), which is not the case for unusual classes and sensor modalities, and increasingly difficult to produce by user annotation for complex structures. 

% Weakly supervised
\myparagraph{Weakly supervised}
In light of this, there has been a push to reduce the requirement of labels. For example, when 3D input data is available, \cite{griffiths2020finding} show that 3D object detection can be performed with significantly less training data through multi-network training. Similarly, \cite{hou2021exploring} also achieve competitive performance with only 1\% of the available training data through pre-training procedures. More recently, \cite{ren20213d} achieve 3D object detection without any spatial labels, requiring only annotation of objects present. When 2D data is available \cite{qin2020weakly} show the combination of an image-based CNN teacher with a 3D object proposal network can be trained without 3D labels. \cite{qin2021monogrnet} also show through careful network design 2D labels can be effective for 3D object detection.

\myparagraph{Self supervised}
Another popular form of supervision for the task of 2D to 3D, that does not require any human annotation or depth sensor is to exploit video or stereo video \cite{godard2017unsupervised, godard2019digging, zhou2017unsupervised}.
Furthermore, with progress in differentiable rendering \cite{loper2014opendr, chen2019learning, li2018differentiable}, analysis-by-synthesis approaches can be used in gradient-based learning. This has led to a number of research works proposing render-and-compare as a replacement for supervision. \cite{kundu20183d} learn both 3D shape and pose with only 2D instance masks. \cite{zakharov2020autolabeling} achieve the same through the additional use of LiDAR data. \cite{beker2020monocular} build on these works, however, remove the need for depth supervision by using an off-the-shelf monocular object detector. Other applications include 6-DOF pose estimation for single objects (equivalent to the LineMOD experiment in Sec. \ref{sec:linemod}).

\myparagraph{Radiance fields}
Implicit representations \cite{sitzmann2019scene,mildenhall2020nerf,bemana2020xfields,niemeyer2020differentiable} fall into this class of supervision by multiple images of one scene and likely owe their superb quality to the fact they do not generalize across multiple objects or even scenes, only across view, light, or time. Recent work has extended multi-view radiance fields to 3D object discovery on simply scenes; however, objects require unrealistic material properties. Furthermore, while our approach supports learning from $n$-view data in the same way that it allows testing on $n-$view image sets, we are not limited to that and generalize to settings where a multi-view (or more general, one-scene-multiple-condition) supervision signal is not available. 

\mysection{Differentiable Rendering with Curiosity}{OurApproach}

Input to our network \network is a 2D image \image and output is a scene code \sceneCode.
The network is a composition $\network(\image)=\rendering(\encoder(\image))$ of an encoder \encoder and a DR \rendering.
The encoder is a CNN that has learnable parameters \parameters to first map to a latent code \latentCode and an MLP to map this code to an explicit scene code \sceneCode.
The DR has no learnable parameters, and additionally, requires only a geometric representation of desired objects.
At test time the rendering (and therefore geometric representation) is not necessary for tasks only requiring the scene code \sceneCode (\eg 3D object detection).
However, for re-synthesis tasks, we can directly modify individual nodes \sceneCode$_i$ of the scene code \sceneCode to manipulate specific elements on the input image, such as shape, color, and orientation.

\myfigure{Simple}{Our architecture is simple. First, a learned mapping encodes an RGB image to scene parameters and second, a fixed differentiable renderer maps the predicted parameters to an image.
Loss is between input and output images, and at deployment we are interested in the scene code for downstream tasks.
Training this analysis-by-synthesis loss without a critic influencing the generated data distribution will almost never converge due to ambiguity (see \refFig{Concept}).}

Common analysis-by-synthesis would minimize $$\loss(\parameters)=||\network_\parameters(\image)-\image||_2.$$
We do not seek to minimize another criterion and would argue there is no need to do so.

The problem is that for almost every possible parameter \parameters, the gradients $\frac{\mathrm d\loss}{\mathrm d\parameters}$ will make the result worse due to the countless ambiguities in image formation.
Consider for example, a badly placed circle resulting from a choice of parameters $\parameters_1$ in a 2D problem of position $\parameters_\mathrm x$ and radius $\parameters_\mathrm r$.
No small change to its parameters will improve the L2 error with respect to the input image.
Instead, gradients will point towards a solution where the sphere just gets smaller or hides in the background color (if this is part of the scene vector). 
% Under image-difference loss, whatever it takes to make no difference is better than trying to do the right thing.
However, there are positions $\parameters_2$, where gradients indeed lead to improvement, \ie when the scene code places the re-rendering of the sphere ``closer'' (this is not limited to space, but happens in the high-dimensional scene code space) to the correct solution.
How can we distinguish those sub-spaces of the solution space from others and gear the optimization to follow these?

The key observation is that the re-rendered solutions from invalid local minima ---in a reverse Anna Karenina principle fashion, according to which all good solutions would be similar and all bad ones unique--- will result in a distribution of re-rendered scenes that are all very similar, \eg all spheres would come out small.
And this distribution is different from the input distribution.
So, all that is needed to push the distributions to become similar, is a second network to compare the data and generated distribution (a critic, in a GAN):
\[
\loss(\parameters)=||\network_\parameters(\image)-\image||_2+\weight\cdot\critic(\network_\parameters(I)).
\]

Our curiosity term is versatile and can influence our scene parameterization network in various ways, depending on the task.
For example, if $\sceneCode \in \mathbb R^6$ our critic will enforce the distribution to match both position ($x,y,z$) and color ($r,g,b$) of the input distribution.
On the other hand, if $\sceneCode \in \mathbb R^6$, but all objects in the input images are the same color, the critic would constrain the parameterization network to always produce predictions with a single fixed object color.

In practice our critic is a simple fully-convolutional CNN encoder $\critic_{\theta}: \image \in \mathbb{R}^{h\times w\times 3} \rightarrow \mathbb{R} ^{1 \in [0, 1]}$.

\mysubsection{Differentiable Rendering with Confidence}{Confidence}
In the above formulation, analysis-by-synthesis with curiosity works well if we have a fixed-size parameter vector, however, would struggle when the number of objects in the scene is unknown.
In the 2D or 3D object detection literature, this is routinely resolved by working with a number of proposals which all carry a confidence \cite{qi2019deep}.
This is then used to suppress objects with a low confidence (visible to the loss or not).
However, without scene-level labels, we not only lack information of object parameters, but also the object count for a given scene.

\myfigure{Confidence}{Confidence in the ideal and real case, and how it is learned. Top row shows a case of two spheres in a scene, bottom row shows a case of one sphere in scene.}

Consider a network which outputs a confidence for every object as part of the scene code \sceneCode.
Ideally, it would be 0 for proposals where no object is present, and 1 for correct proposals, as seen in \refFig{Confidence}, left, but in practice is a fraction as seen in \refFig{Confidence}, middle.
This confidence has to affect how the scene is rendered if we want it to be learned.
Simply turning an object on and off based on any threshold is not differentiable.
Instead, we suggest a soft rendering ($\rendering_\mathrm c$) which enables the network to learn to hide redundant predictions.

The easiest way to implement a soft rendering would be to modify the objects transparency in the DR. However, as the DR we use in our experiments \cite{li2018differentiable} does not support transparency, we emulate the behavior as follows: we first render the scene without any objects and then with every object in isolation, all with an alpha channel.
These images are then composed back-to-front using confidence as alpha, which is a differentiable operation itself.
This is correct for first-hit rays, but incorrect for the compositing of higher-order shading (a confident object casting no shadow at one pixel, might override the correct shadow of another confident object).
In the future we expect DRs to support transparency, allowing us to benefit from guidance by shadow and global illumination for multiple objects.
Note that all results where the object count is known, do have correct unbiased image synthesis and benefit from shadow, reflection, etc. informing the loss.

\myfigure{Datasets}{Samples from the different datasets we study.}

\mysection{Results}{Results}
We will first analyze our approach on synthetic scenes (\refSec{Analysis}) before applying it to real photos (\refSec{Real}). Finally, we evaluate our approach on the popular LineMOD dataset for 6-DOF pose estimation of single objects.

\mysubsection{Analysis}{Analysis}

\myparagraph{Tasks}
We consider a synthetic dataset of renderings of 3D scenes with different parameters (\refFig{Datasets} and \refTbl{Worlds}).

\dataset{Circles} (\iconCircles) is a single red 2D circle of constant radius in front of a black background.
\dataset{Sphere} (\iconSpheres) is a 3D world with three spheres of varying color.
Objects in this task and all following are placed via rejection sampling such as to not intersect.
\dataset{Chairs} (\iconChairs) is a 3D world with five chairs placed on the ground plane at different positions and orientations.
\dataset{Varied} (\iconVaried) comprises of a varying number of between 2 and 5 spheres of random position and color.
This is the first task where confidence-based rendering (\refSec{Confidence}) is used.
\dataset{Multi} (\iconMulti) has eight objects: four sets of two letters ($\mathtt{WXYZ}$), with random colors, placed on the ground plane at random positions and orientations.
\dataset{Light} (\iconLight) is a 3D world with a varying number of different objects in it (capsules, boxes, cones), all at random positions and orientations on a ground plane and illuminated by a single changing ``sunlight'' illumination, that is part of the scene description.

For every task we consider a set of 2000 images, labeled with the scene parameters (hidden to our training) with 500 images for validation and 500 images for testing.
We will later consider fractions of this supervision.

\begin{table}
    \caption{Latent scene code structure.}
    \vspace{.1cm}
    \setlength{\tabcolsep}{0.13cm}
    \centering
    \begin{tabular}{rrrrccccccc}
        \toprule
        Dataset&
        \rotatebox[origin=c]{90}{DoF}&
        \rotatebox[origin=c]{90}{From $n$}&
        \rotatebox[origin=c]{90}{To $n$}&
        \rotatebox[origin=c]{90}{Real}&
        \rotatebox[origin=c]{90}{3D}&
        \rotatebox[origin=c]{90}{Position}&
        \rotatebox[origin=c]{90}{Color}&
        \rotatebox[origin=c]{90}{Rotate}&
        \rotatebox[origin=c]{90}{Shape}&
        \rotatebox[origin=c]{90}{Light}\\
        \midrule
        \dataset{Circles \iconCircles}&
        2&
        1&
        1&
        \xmark&
        \xmark&
        \cmark&
        \xmark&
        \xmark&
        \xmark&
        \xmark
        \\
        \dataset{Sphere \iconSpheres}&
        6&
        3&
        3&
        \xmark&
        \cmark&
        \cmark&
        \cmark&
        \xmark&
        \xmark&
        \xmark
        \\
        \dataset{Chairs \iconChairs}&
        5&
        5&
        5&
        \xmark&
        \cmark&
        \cmark&
        \xmark&
        \cmark&
        \xmark&
        \xmark
        \\
        \dataset{Varied \iconVaried}&
        7&
        2&
        5&
        \xmark&
        \cmark&
        \cmark&
        \cmark&
        \xmark&
        \xmark&
        \xmark
        \\
        \dataset{Multi \iconMulti}&
        9&
        8&
        8&
        \xmark&
        \cmark&
        \cmark&
        \cmark&
        \cmark&
        \cmark&
        \xmark
        \\
        \dataset{Light \iconLight}&
        11&
        2&
        10&
        \xmark&
        \cmark&
        \cmark&
        \cmark&
        \cmark&
        \cmark&
        \cmark
        \\
        \midrule
        \dataset{Sphere Real}&
        6&
        3&
        3&
        \cmark&
        \cmark&
        \cmark&
        \cmark&
        \xmark&
        \xmark&
        \xmark   \\     
        \dataset{Bunny}&
        6&
        1&
        1&
        \cmark&
        \cmark&
        \cmark&
        \cmark&
        \cmark&
        \xmark&
        \xmark   \\     
        \dataset{Shapes}&
        6&
        5&
        12&
        \cmark&
        \cmark&
        \cmark&
        \cmark&
        \cmark&
        \cmark&
        \xmark   \\     
        \bottomrule
    \end{tabular}
    \label{tbl:Worlds}
\end{table}

\myparagraph{Methods}
We consider three methods:
The first is a hypothetical baseline that has access to the ground truth scene parameters learned with a supervised loss.
We study three variants of this method which use 5\% (\nameInColor{Super5}), 10\% (\nameInColor{Super10}) and 100\% of the supervision.
While our method is trained from images alone, we have to define a supervised loss for this baseline.
Direct $\mathcal L_2$ between scene parameters cannot work, as it implies object order.
Direct application of a Chamfer loss based on position is not practical as it would ignore all non-positional attributes, and it would not be clear how to handle global attributes like light direction.
Hence, we first compute the optimal assignment $P$ between $n_\mathrm o$ objects according to only position $\mathbf x$, and then a $\mathbf w$-weighted norm between the scene object attributes $\mathbf y$, including position, and the global attributes $\mathbf z$ of scene $A$ and scene $B$, paired by $P$, as in
\begin{align}
\mathcal L(A,B)&=
\sum_{i=1}^{n_\mathrm o}
\sum_{j=1}^{n_\mathrm p}
w_j
||
\mathbf y^A_{i,j}-
\mathbf y^B_{P_i,j}
||
+
||
\mathbf z^A-
\mathbf z^B
||
,
\nonumber
\\
\text{with \qquad }
P&=\argmin{Q}
\sum_{i=1}^{n_\mathrm o}
||
\mathbf x^A_i-
\mathbf x^B_{Q_i}
||.
\label{eq:Metric}
\end{align}

\nameInColor{NonCur} is a basic self-supervised method, without our curiosity term (L2 loss only) while \nameInColor{Our} is our full method (L2 and critic).

All methods use the same architecture, and only differ by their supervision.
Input is a 3-channel colored image with size 128$\times$128 image, that is reduced to a single latent code $\latentCode$ of 64 dimensions in 7 steps (with Batch norm and ReLU).
An MLP with three non-linear projection layers maps the latent code $\latentCode$ into a scene code $\sceneCode$ of a size that depends on the task (between 2 and 11 values, see \refTbl{Worlds}) to control the DR (\refFig{Simple}).
The first layer of the MLP is a shared trunk, while the second and third layers produce the respective scene parameters used in independent branches for each group: one for 3D position, 2D rotation, 3D color, 1D confidence and 2D light.

\myparagraph{Metrics}
We apply two different metrics: \refEq{Metric} on the resulting scene parameters and image error, where we render the scene from a novel viewpoint (to avoid single image scale ambiguity) with the estimated and the ground truth scene parameters and compare the images using Structural Dissimilarity (DSSIM).
We report numbers as a ratio relative to the fully supervised method on that task (not shown; it always is $1.00$), as this is the upper bound of our relatively simple detection network.

\begin{table}[htb!]
    \setlength{\tabcolsep}{0.05cm}
    \caption{Different methods and supervisions \emph{(columns)} according to different metrics for different data sets \emph{(rows)}.
    The plot shows the 100\,\% supervision-reference as a thick line.}
    \label{tbl:Results}
    \vspace{0.1cm}
    \centering
    \begin{tabular}{c rrrrrrrrrrr}
        \toprule
        &
        \multicolumn2c{\nameInColor{Super5}}&
        \multicolumn2c{\nameInColor{Super10}}&
        \multicolumn2c{\nameInColor{NonCur}}&
        \multicolumn2c{\nameInColor{Our}}\\
        \cmidrule(lr){2-3}
        \cmidrule(lr){4-5}
        \cmidrule(lr){6-7}
        \cmidrule(lr){8-9}
        &
        \multicolumn1c{\footnotesize{Img}}&
        \multicolumn1c{\footnotesize{Para}}&
        \multicolumn1c{\footnotesize{Img}}&
        \multicolumn1c{\footnotesize{Para}}&
        \multicolumn1c{\footnotesize{Img}}&
        \multicolumn1c{\footnotesize{Para}}&
        \multicolumn1c{\footnotesize{Img}}&
        \multicolumn1c{\footnotesize{Para}}
        \\
        \midrule
\dataset{Circles}\iconCircles & 2.31 & 8.69 & 1.61 & 3.51 & 5.08 & 3621.83 & 0.94 & 1.11 \\
\dataset{Spheres}\iconSpheres & 1.67 & 2.33 & 1.59 & 2.29 & 1.68 & 8.97 & 1.19 & 1.87 \\
\dataset{Chairs}\iconChairs & 1.06 & 3.48 & 1.04 & 1.54 & 2.23 & 51.78 & 1.03 & 1.40 \\
\dataset{Varied}\iconVaried & 1.48 & 10.61 & 1.27 & 5.84 & 2.28 & 22.47 & 1.25 & 5.04 \\
\dataset{Multi}\iconMulti & 1.73 & 2.59 & 1.28 & 1.78 & 2.11 & 18.98 & 1.14 & 1.51 \\
\dataset{Light}\iconLight & 2.55 & 5.44 & 1.24 & 2.53 & 3.09 & 18.65 & 1.12 & 2.05 \\
\bottomrule
    \end{tabular}
    \includegraphics[width=\linewidth]{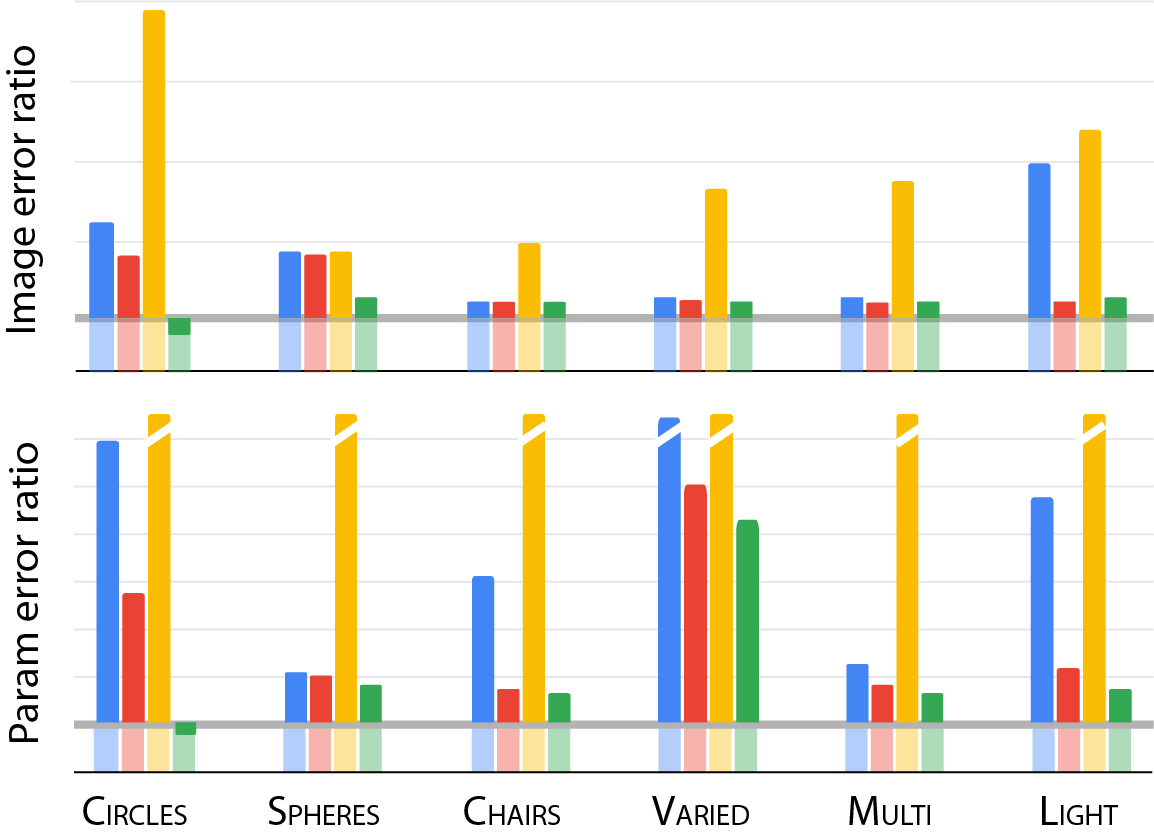}%
\end{table}

\begin{table*}[]
   \setlength{\tabcolsep}{0.08cm}
    \centering
    \caption{Details of the per-parameter error of different methods in ratios with respect to a supervised reference.
    The ratio indicates, by what factor a method is worse, compared to a 100\% supervision-baseline.}
    \label{tbl:Details}
    \vspace{.1cm}
\begin{tabular}{crrrrrrrrrrrrrrrrrrrr}
\toprule
&
\multicolumn{4}{c}{Position (m)}&
\multicolumn{4}{c}{Color (MSE)}&
\multicolumn{4}{c}{Rotation (Deg)}&
\multicolumn{4}{c}{Confidence (MSE)}&
\multicolumn{4}{c}{Direction (deg)}
\\
\cmidrule(lr){2-5}
\cmidrule(lr){6-9}
\cmidrule(lr){10-13}
\cmidrule(lr){14-17}
\cmidrule(lr){18-21}
&
\shortNameInColor{Super5}{5\%}&
\shortNameInColor{Super10}{10\%}&
\shortNameInColor{NonCur}{NC}&
\shortNameInColor{Our}{O}&
\shortNameInColor{Super5}{5\%}&
\shortNameInColor{Super10}{10\%}&
\shortNameInColor{NonCur}{NC}&
\shortNameInColor{Our}{O}&
\shortNameInColor{Super5}{5\%}&
\shortNameInColor{Super10}{10\%}&
\shortNameInColor{NonCur}{NC}&
\shortNameInColor{Our}{O}&
\shortNameInColor{Super5}{5\%}&
\shortNameInColor{Super10}{10\%}&
\shortNameInColor{NonCur}{NC}&
\shortNameInColor{Our}{O}&
\shortNameInColor{Super5}{5\%}&
\shortNameInColor{Super10}{10\%}&
\shortNameInColor{NonCur}{NC}&
\shortNameInColor{Our}{O}
\\
\midrule
\dataset{Circle} & 8.69 & 3.51 & 362 & 1.11 & \nacell & \nacell & \nacell & \nacell & \nacell & \nacell & \nacell & \nacell & \nacell & \nacell & \nacell & \nacell & \nacell & \nacell & \nacell & \nacell \\
\dataset{Sphere} & 5.14 & 3.79 & 18.77 & 2.88 & 2.36 & 1.67 & 4.94 & 1.46 & \nacell & \nacell & \nacell & \nacell & \nacell & \nacell & \nacell & \nacell & \nacell & \nacell & \nacell & \nacell \\
\dataset{Chair} & 2.89 & 2.22 & 206 & 1.19 & \nacell & \nacell & \nacell & \nacell & 3.64 & 1.37 & 12.1 & 1.45 & \nacell & \nacell & \nacell & \nacell & \nacell & \nacell & \nacell & \nacell \\
\dataset{Varied} & 5.29 & 2.65 & 4.91 & 1.25 & 2.50 & 1.72 & 4.51 & 1.74 & \nacell & \nacell & \nacell & \nacell & 20.9 & 11.3 & 48.3 & 10.1 & \nacell & \nacell & \nacell & \nacell \\
\dataset{Multi} & 2.76 & 2.20 & 28.43 & 1.41 & 2.37 & 1.55 & 4.43 & 1.69 & 2.52 & 1.32 & 16.5 & 2.95 & \nacell & \nacell & \nacell & \nacell & \nacell & \nacell & \nacell & \nacell \\
\dataset{Light} & 4.01 & 2.18 & 41.57 & 1.39 & 2.06 & 1.49 & 2.95 & 1.55 & 1.92 & 1.51 & 14.0 & 1.24 & 14.8 & 5.33 & 20.2 & 4.47 & 1.62 & 1.32 & 1.54 & 1.14 \\
\bottomrule
\end{tabular}
\end{table*}

\myparagraph{Findings}
We computed the result of all methods on all synthetic datasets for all metrics (\refTbl{Results}).
The main finding is, that \nameInColor{Our} method with curiosity performs better than \nameInColor{NonCur} which has no curiosity.
This is compared with methods that have supervision (and do not need curiosity) at 5\,\% (\nameInColor{Super5}) and 10\,\% (\nameInColor{Super10}) of the data.
Unsurprisingly, more supervision leads to lower errors.
The performance of full supervision is shown as a thick line in the plot of \refTbl{Results} and matches 1.00.
We see that in most tasks and for several metrics our method can perform similar to the 100\,\%-supervised baseline, while it had no access to scene labels.
In general, the difference in scene parameter error is larger than the one in image error.
In particular, the scene parameter error is too high for \nameInColor{NonCur} to be plotted. 
More surprisingly, and affirmative, differences according to both metrics seem to be diminishing when the task gets more complex, \eg comparing the progression of  \colorIconAndName{Our}{Circles} / \colorIconAndName{Our}{Spheres} / \colorIconAndName{Our}{Chairs} vs.
\colorIconAndName{Super10}{Circles} / \colorIconAndName{Super10}{Spheres} / \colorIconAndName{Super10}{Chairs}, we see that the gap between supervision and no supervision closes, a progression found for both metrics.

\refFig{Results} shows qualitative results for selected tasks.
The parameter vector is visualized as a scene graph in an 3D modeling application.
Note, that only the orientation, positions etc. (as defined in \refTbl{Worlds}) is part of our method's output.

\refTbl{Details} splits the parameter error of all scenes between all applicable classes of scene parameters (position, color, orientation, confidence, and light direction) present in our scenes.
We see the largest error ratio to be found with position, in particular for \nameInColor{NonCur}.
Extracting the color in the simple scenes we use is not a particularly hard task, so we find all methods to not lose much quality from less supervision.
The error ratios for rotation are smaller, probably as when the network already has learned to position the object, getting rotation right is an easier optimization task.
The confidence error ratio is higher for (\colorIconAndName{Our}{Multi}) than for (\colorIconAndName{Our}{Light}), likely as the object geometry between shapes in the latter have a higher variance.
It could be hypothesized that varied illumination also helps understanding, but we did not study this difference.
Finally, (\colorIconAndName{Our}{Light}) contributes little to the overall error, indicating the network has learned light direction from 2D images alone, almost as it had done with supervision. 

\myfigure{Analytic}{Solving an analytic problem with and without curiosity.
Blue points are initial and red points ground truth; grey lines show correspondence.
Optimization is the blue-red trajectory.
}
\myparagraph{Validation experiment.
}
The effect of curiosity can be verified for a very basic analytic 2D problem, finding positions ($x$) and luminance ($l$) of $N$ objects in $N$ images.
We visualize the $N$ solution as a red 2D point set in the position-luminance plane in \refFig{Analytic}.
Starting from random initial guesses (blue 2D points in \refFig{Analytic}, a), optimizing those $N$ problems independently, analysis-by-synthesis will in almost all cases converge (trajectories and blue points in \refFig{Analytic}, a) to a degenerate and incorrect $l=0$.
Consider a world where $x$ and $l$ follow a uniform random distribution.
Deviation of the set of resulting $x,l$ pairs from a uniform distribution can be measured by computing discrepancy in closed form.
This deviation provides an idealized measure of curiosity.
Adding this term forces the distributions of solutions to be uniform over the $x,l$ plane (\refFig{Analytic}, b), finds the correct solution for almost all problems jointly. For more details see the appendix (Sec. \ref{appendix}).

\mysubsection{Real world data}{Real}
While we have initially evaluated our approach in a virtual setting, which is suitable for instrumentation as all scene parameters are known, we now take it to a real setting, where parameters are unknown.

\myparagraph{Data}
For training data, we capture sequences of objects in front of a neutral background using standard cameras.
For evaluation, we took several photos from multiple directions, allowing us to perform Structure-from-Motion to retrieve camera matrices (see appendix (Sec. \ref{appendix}) for more details).
We captured the datasets \dataset{Sphere Real}, \dataset{Bunny} and \dataset{Shapes} (\refFig{RealDatasets}).
In both, the task is to learn to regress 3D position, orientation and light from an image as summarized in the lower part of \refTbl{Worlds}.

\myfigure{RealDatasets}{Three \dataset{Cube} and \dataset{Bunny} samples from the real capture datasets.
Features like sensor noise, motion blur, geometric details and depth-of-field are not reproduced by our DR, while still the method can be trained.}

\myparagraph{Methods}
The data available only allows studying \nameInColor{NonCur} in comparison to \nameInColor{Our} as no supervision parameter values are known. To close the domain gap, we blur the rendering and camera image by a 9$\times$9-pixel Gaussian filter to help remove camera noise.

\myparagraph{Metric}
Sets of images of the test scene from multiple views of known relative pose allow to test the reprojection task, as for synthetic data using DSSIM.
The object parameters remain unknown and cannot be compared to anything for the real experiment.
Consequently, results also cannot be reported as ratio relative to this supervision signal.
Further, for the above-mentioned reasons (camera noise, MB, DoF), there is a domain gap between any of our novel-view images and the reference.
To approximately quantify what differences might be if that domain gap would not be there, we additionally report the DSSIM error for the blurred version of both images.
While the error can be quantified, the reader might get a better impression from looking at the actual re-synthesized images in \refFig{Results}.

\mycfigure{Results}{
Results of our approach.
Each row is a different world / task.
Note how there was no other supervision than sample images.}

\myparagraph{Findings}
We find that our method is able to recover sensible parameters when trained on only real RGB images.
\refTbl{RealResults} quantifies this. We see that analysis-by-synthesis training without curiosity (\nameInColor{NonCur}) results in a higher image error than \nameInColor{Our}.
This corresponds to qualitative results in \refFig{Results}.

\begin{table}
    \caption{Results on real photos for different datasets \textbf{(columns)} for different methods \textbf{(rows)}.
    Image error is in DSSIM, not ratio, as no supervised baseline on photos exist.}
    \label{tbl:RealResults}
    \setlength{\tabcolsep}{0.16cm}
    \vspace{.1cm}
    \centering
    \begin{tabular}{rrrrrrr}
        \toprule
        &
        \multicolumn2c{\dataset{Sphere Real}}&
        \multicolumn2c{\dataset{Bunny}}&
        \multicolumn2c{\dataset{Shapes}}
        \\
        \cmidrule(lr){2-3}
        \cmidrule(lr){4-5}
        \cmidrule(lr){6-7}
        &
        Sharp&
        Blur&
        Sharp&
        Blur&
        Sharp&
        Blur
        \\
        \midrule
        \nameInColor{NonCur}&
        0.48&
        0.41&
        0.46&
        0.43&
        0.54&
        0.46
        \\
        \nameInColor{Our}&
        0.33&
        0.25&
        0.36&
        0.31&
        0.37&
        0.29
        \\
        \bottomrule
    \end{tabular}
\end{table}

\mysubsection{LineMOD}{linemod}

\mycfigure{LineMod}{Visualization of (\textbf{top, green}) ground truth boxes and (\textbf{bottom, red}) our network predictions on LineMOD test data.}

\begin{table*}[b]
    \setlength{\tabcolsep}{0.15cm}
    \small
    \centering
    \caption{Results from LineMOD experiment. All results denote the ADD object recall metric proposed by \cite{hinterstoisser2012model}. 'Lbl\textsubscript{R}' and 'Lbl \textsubscript{S}' indicates whether real or synthetic pose or mask labels were used respectively. Note that our approach and \cite{sundermeyer2020augmented} use neither.}
    \begin{tabular}{lccrrrrrrrrrrrrrr}
        \toprule
        \multicolumn1c{Method}&
        \multicolumn1c{Lbl\textsubscript{R}}&
        \multicolumn1c{Lbl\textsubscript{S}}&
        \multicolumn1c{Avg.}&
        Ape&
        B.vise&
        Cam&
        Can&
        Cat&
        Drill&
        Duck&
        Eggb.&
        Glue&
        Pun.&
        Iron&
        Lamp&
        Phone \\
        
        \midrule
        Yolo6d \cite{tekin2018real} & \cmark & \xmark & 56.0 & 21.6 & 81.8 & 36.6 & 68.8 & 41.8 & 63.5 & 27.2 & 69.6 & 80.0 & 42.6 & 75.0 & 71.1 & 47.7\\
        Pix2Pose \cite{park2019pix2pose} & \cmark & \xmark & 72.4 & 58.1 & 91.0 & 60.9 & 84.4 & 65.0 & 76.3 & 43.8 & 96.8 & 79.4 & 74.8 & 83.4 & 82.0 & 45.0\\
        Kehl et al. \cite{kehl2017ssd} & \cmark& \xmark & 92.9 & \bestSupervised{98.1} & 94.8 & 93.4 & 82.6 & \bestSupervised{98.1} & 96.5 & 97.9 & \bestSupervised{100.0} & 74.1 & \bestSupervised{97.9} & 91.0 & 98.2 & 84.9\\
        EfficientPose \cite{bukschat2020efficientpose} & \cmark & \xmark & \bestSupervised{97.4} & 87.7 & \bestSupervised{99.7} & \bestSupervised{97.9} & \bestSupervised{98.5} & 98.0 & \bestSupervised{99.9} & 91.0 & \bestSupervised{100.0} & \bestSupervised{100.0} & 95.2 & \bestSupervised{99.7} & \bestSupervised{100.0} & \bestSupervised{98.0}\\
        Our (label)& \cmark& \xmark & 78.2 & 71.5 & 93.4 & 70.2 & 71.8 & 76.3 & 73.8 & 65.4 & 90.5 & 71.5 & 86.2 & 81.4 & 86.2 & 78.5\\
        \midrule
        Yolo6d \cite{tekin2018real} & \xmark & \cmark & 21.4 & 16.1 & 33.9 & 2.9 & 21.0 & 27.1 & 24.7 & 20.2 & 2.3 & 15.0 & 15.4 & 57.6 & 26.8 & 14.8\\
        Pix2Pose \cite{park2019pix2pose} & \xmark& \cmark& 11.3 & 3.6 & 4.0 & 0.0 & 16.6 & 20.2 & 29.0 & 0.2 & 0.0 & 7.3 & 2.5 & 1.8 & 30.2 & 31.9\\
        Kehl et al. \cite{kehl2017ssd} & \xmark& \cmark& 2.4 & 0.0 & 0.2 & 0.4 & 1.4 & 0.5 & 2.6 & 0.0 & 8.9 & 0.0 & 0.3 & 8.9 & 8.2 & 0.2\\
        SynPo-Net \cite{su2021synpo} & \xmark & \cmark & 44.1 & 23.1 & \bestUnsupervised{75.2} & 6.7 & \bestUnsupervised{65.1} & 36.2 & 53.7 & 19.5 & 3.9 & 41.8 & 21.1 & \bestUnsupervised{85.1} & 78.7 & 63.6\\
        Implicit3d \cite{sundermeyer2020augmented} & \xmark & \xmark & 32.7 & 4.2 & 22.9 & 32.9 & 37.0 & 18.7 & 24.8 & 5.9 & \bestUnsupervised{81.0} & 46.7 & 18.2 & 35.1 & 61.2 & 36.3\\
        Our \nameInColor{NonCur}& \xmark& \xmark& 1.6 & 2.6 & 1.5 & 0.4 & 0.9 & 1.1 & 1.5 & 0.0 & 5.2 & 1.5 & 3.2 & 0.0 & 0.0 & 2.3\\
        \nameInColor{Our} & \xmark& \xmark& \bestUnsupervised{60.9} & \bestUnsupervised{50.2} & 54.6 & \bestUnsupervised{63.5} & 58.1 & \bestUnsupervised{42.6} & \bestUnsupervised{58.4} & \bestUnsupervised{38.6} & 68.0 & \bestUnsupervised{56.7} & \bestUnsupervised{68.2} & 78.3 & \bestUnsupervised{79.4} & \bestUnsupervised{75.3}\\
         \bottomrule
    \end{tabular}
    \label{tab:linemod-results}
\end{table*}

This task entails regressing a 6-DOF pose of a known object in images \cite{hinterstoisser2012model}. Unfortunately, we do not have access to scenes without the objects present to generate training data. As a result, to achieve the optimum L2 and critic loss our network could simply render the objects out of the cameras view. To account for this, we take a similar approach to \cite{su2021synpo} whereby we generate synthetic images by rendering 3D models over 2D background images. We first collect backgrounds by downloading 500 images returned by a Google search of the term ``clutter''. For each class we generate 10,000 training images where the respective mesh has a random position and orientation and each scene a random environment light. Unlike \cite{su2021synpo}, we do not access synthetic labels during training.

We use the same network architecture proposed in Sec. \ref{sec:OurApproach}. However, our scene code $ \sceneCode \in \mathbb R^{12}$ learns position $p \in \mathbb R^3$, rotation $R \in \mathbb R^6$ (as proposed by \cite{zhou2019continuity}) and environment light $L \in \mathbb R^3$. We use data loading scheduling to train our model. First, we train only on synthetic images to learn to make predictions in the camera view. We then incrementally add real training images (without accessing labels).

Our method on average out-performs all methods in Tbl. \ref{tab:linemod-results} where either no labels or synthetic labels are used. We perform competitively compared to supervised approaches, with a 20\% reduction from ours supervised (the upper bound of our detector). Qualitative results are shown in \refFig{LineMod}.

\mysection{Conclusion and Future Work}{Conclusion}
We have demonstrated a pipeline combining DR and a curiosity term, that allows to learn an explicit and interpretable 3D scene parameterization of a single image from a set of unlabeled 2D images and a geometric representation of the objects we wish to find.
Different from other approaches, we are not limited to simple or single objects and can represent global properties such as light.
Direct optimization under a L2-like loss function leads to many ambiguities, that can be overcome by what looks like a GAN-like critic at first but serves an entirely different purpose: preventing the network from following always the same easy-reward gradients that lead to unusable minima. With this simple addition we show a relatively basic network, without labels, can perform competitively with a fully supervised counterpart.

A limitation to our approach is the requirement of a geometric representation. In future work we propose such a template, if represented as a parameterizable model, such as a deep Signed Distance Function (SDF) \cite{park2019deepsdf} from a strong prior, can be included in the scene code and learnt during training. \cite{beker2020monocular} demonstrate such an approach to be effective.

\clearpage
{\small
\bibliographystyle{ieee_fullname}
\bibliography{egbib}
}

\clearpage

\section{Appendix}\label{appendix}

\subsection{Network architecture}

In this section we give further details of our neural network architectures. For simplicity we ignore the batch dimension.

Our encoder is an approximation of the AlexNet architecture. Input is a 3-channel colored image with size $128\times128$. At each layer we reduce the image dimensions using strided convolutions, avoiding the need for max pooling operations. We use a Rectified Linear Unit (ReLU) for all non-linear activations. Whilst a deeper network, or one with more modern operations (e.g. Residual connections), would likely have helped overall performance, we aimed to keep are architecture as simple as possible to explore the relative benefits of our proposed learning approach.

\begin{table}[h!]
    \caption{AlexNet encoder architecture details.} \label{tbl:encoder}
    \vspace{.1cm}
    \setlength{\tabcolsep}{0.15cm}
    \centering
    \begin{tabular}{ c c c c c c} 
         \hline
         & Layer & Out Size & Batch Norm & Activation \\
         \hline
         C1 & Conv 2D & [64, 30, 30] & True & ReLU \\ 
         C2 & Conv 2D & [192, 14, 14] & True & ReLU \\ 
         C3 & Conv 2D & [384, 7, 7] & True & ReLU \\ 
         C4 & Conv 2D & [256, 4, 4] & True & ReLU \\ 
         C5 & Conv 2D & [64, 1, 1] & False & ReLU \\ 
         \hline
         \multicolumn{5}{r}{Parameters: 2,029,056}

    \end{tabular}
    \includegraphics[width=\linewidth]{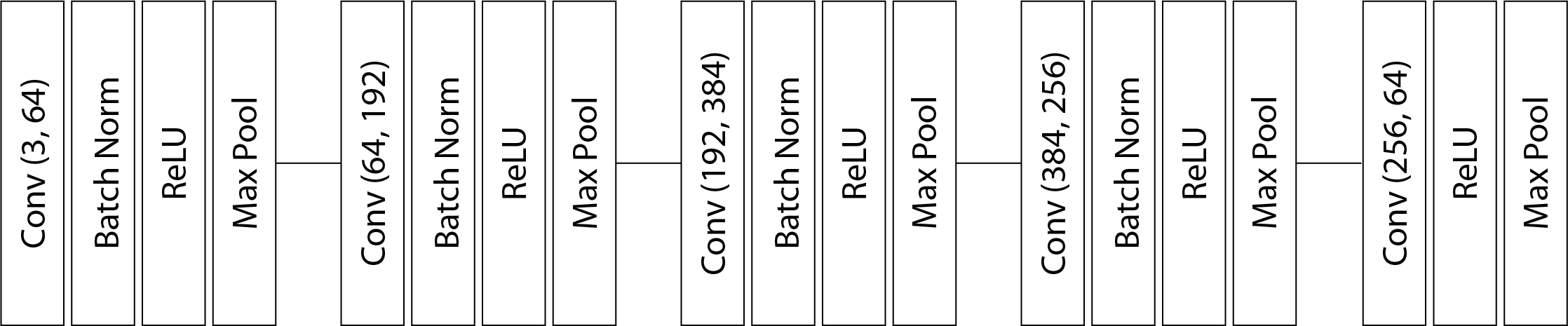}
\end{table}

Our critic architecture is fully convolutional, and therefore contains no fully connected layers. Unlike the image encoder we find Leaky ReLU activation functions to be more effective. The critic also contains no max pooling layers, using convolutions with either stride=2 or stride=4 to reduce the image size. Input is a 3-channel colored image with size $128\times128$. In particular with the critic, we found normalization to be important. Whilst both Instance Normalization and Batch Normalization proved to be effective, all results in the paper were obtained using Batch Normalization.

\begin{table}[h!]
    \caption{Critic architecture details} \label{tbl:critic}
    \vspace{.1cm}
    \setlength{\tabcolsep}{0.15cm}
    \centering
    \begin{tabular}{ c c c c c c} 
     \hline
     & Layer & Out Size & Batch Norm & Activation \\
     \hline
     C1 & Conv 2D & [64, 30, 30] & True & Leaky ReLU \\ 
     C2 & Conv 2D & [192, 7, 7] & True & Leaky ReLU \\ 
     C3 & Conv 2D & [384, 4, 4] & True & Leaky ReLU \\ 
     C4 & Conv 2D & [256, 2, 2] & True & Leaky ReLU \\ 
     C5 & Conv 2D & [64, 1, 1] & False & Sigmoid \\ 
     \hline
     \multicolumn{5}{r}{Parameters: 1,883,713}
    \end{tabular}
    \includegraphics[width=\linewidth]{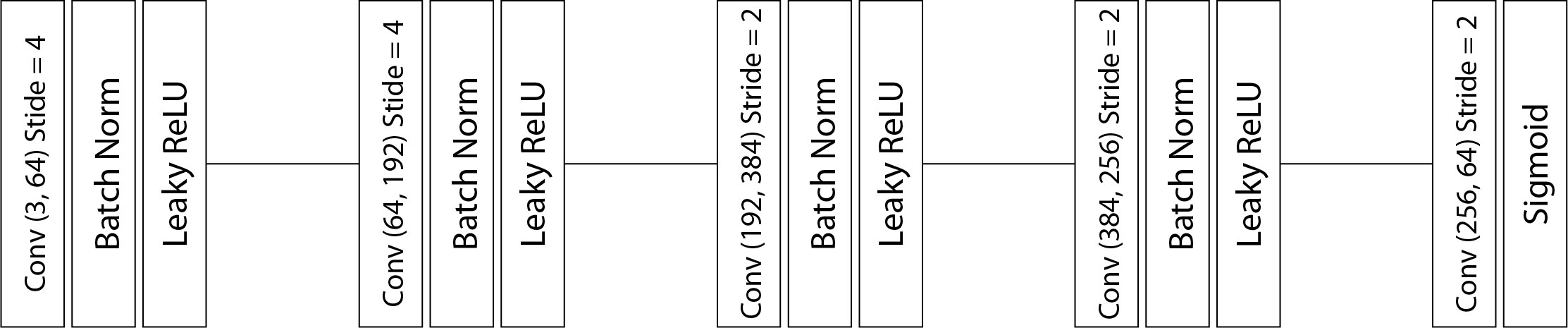}
\end{table}

The parameter prediction MLP contains a single shared trunk layer followed by respective parameter branches. No weights are shared on any branch layers. After the shared layers (trunk) we resize the vector to size $n\times d$ where $n$ is the number of object predictions and $d$ number of feature dimensions. We find this effective as it enables each prediction to have its own independent feature transformation. $\mathtt{Center}$ predicts the per-object translation offset in camera space. $\mathtt{RGB}$ predicts an per-object RGB color. $\mathtt{Confidence}$ predicts a per-object confidence which is subsequently used as an opacity value for differentiable rendering with varying objects. $\mathtt{Light}$ predicts a per-scene light direction along a hemisphere above the scene. 

For experiments where a canonical mesh orientation, and ground plane are assumed, $\mathtt{Rotation}$ predicts the per-object $i,j$ rotation vector direction. The final rotation $r$ around the $y$-axis is calculated as $r=\mathrm{arctan2}(j, i)$. For experiments where this is not the case (\textsc{Bunny}, \textsc{LineMOD}) we instead using a 6D rotation representation as proposed by \cite{zhou2019continuity}. We choose this representation as it is continuous unlike other representations (Euler angles, rotation matrix, Quaternions etc.), which is more suitable for gradient decent optimizations.

We do not use any normalization in between MLP layers.

\begin{table}[h!]
    \caption{Parameter MLP network architecture.} \label{tbl:mlp}
    % \vspace{.01cm}
    \setlength{\tabcolsep}{0.15cm}
    \centering
    \begin{tabular}{ l c c c} 
         \hline
          & Layer & Out Size & Activation \\
         \hline
         Trunk & Fully Connected & [64, 1] & ReLU \\ 
        \hline
         \multirow{2}{5em}{$\mathtt{Center}$} & Fully Connected & [n, 64] & ReLU \\ 
          & Fully Connected & [n, 3] & Linear \\ 
        \hline
         \multirow{2}{5em}{$\mathtt{Rotation}$} & Fully Connected & [n, 64] & ReLU \\ 
         & Fully Connected & [n, 2] & TanH \\ 
        \hline
         \multirow{2}{5em}{$\mathtt{RGB}$} & Fully Connected & [n, 64] & ReLU \\ 
         & Fully Connected & [n, 3] & Sigmoid \\ 
        \hline
         \multirow{2}{5em}{$\mathtt{Confidence}$} & Fully Connected & [n, 64] & ReLU \\ 
         & Fully Connected & [n, 1] & Sigmoid \\ 
        \hline
         \multirow{2}{5em}{$\mathtt{Light}$} & Fully Connected & [1, 64] & ReLU \\ 
         & Fully Connected & [1] & Linear \\ 
         \hline
         \multicolumn{4}{r}{Parameter range: 4,387 - 48,266}
    \end{tabular}
    \includegraphics[width=\linewidth]{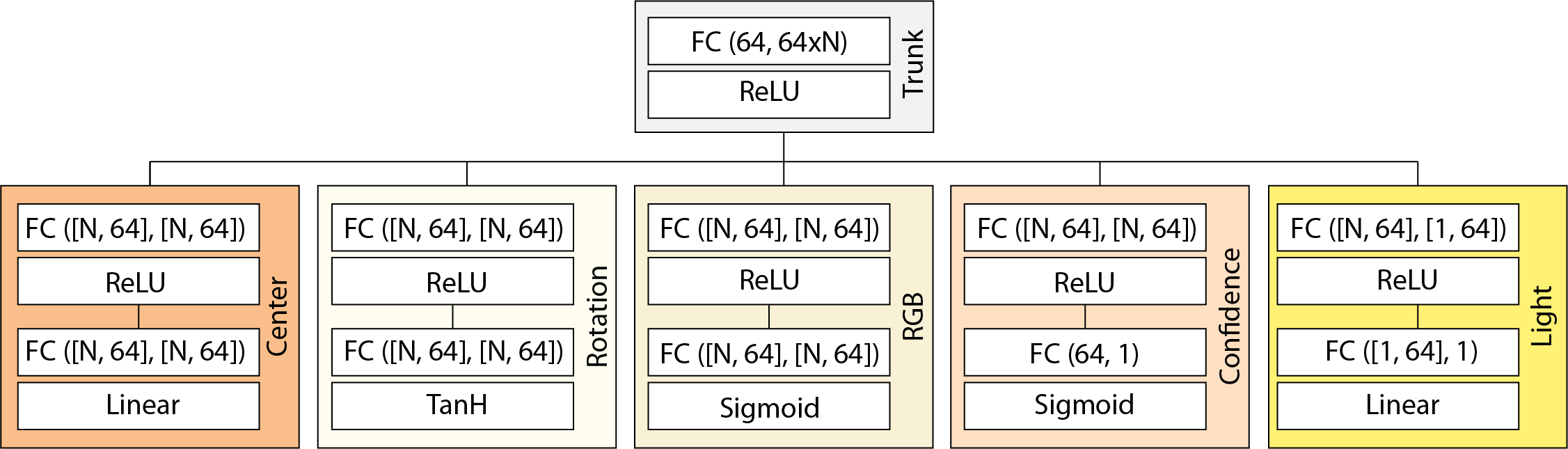}
\end{table}

\subsection{Training details}\label{sec:training_details}
All networks were implemented in the PyTorch v1.7 framework. A list of hyperparameters are shown in Tab. \ref{tab:HyperParams}. We trained all models until a convergence threshold based on the image loss is reached. Due to the nature of generator-critic min-max games, we found the critic loss to be an unreliable indicator of convergence. Network training times took between 12-36 hours on a single Nvidia 2080ti depending on task. To allow for a high batch size we utilize virtual batch sizes to avoid running into memory constraints. We achieve this by running multiple forward passes with a smaller batch size and accumulating the gradients before running a backwards pass through our network.

\begin{table}[]
    \centering
    \begin{tabular}{r|c}
        \hline
         Hyperparameter & Value  \\
         \hline
         Batch size & 128 \\
         Optimizer & Adam ($\beta_1$: 0.9, $\beta_2$: 0.999) \\
         Generator Learning rate & 1e-4 \\
         Critic learning rate & 1e-6 \\
         Image Loss $\lambda$ & 0.01 \\
         Critic loss $\lambda$ & 10 \\
         Gradient L2 clipping & 0.5 \\
    \end{tabular}
    \caption{List of hyperparameters used to train networks.}
    \label{tab:HyperParams}
\end{table}

\subsection{Real data experimental setup}\label{sec:exp_setup}
\mycfigure{RealSetup}{Real dataset acquisition pipeline. Images are first captured from either mobile (\textsc{Bunny}) or fixed (\textsc{Sphere Real, Shapes}) cameras (\textbf{a}). Camera alignment is calculated using SfM (\textbf{b}). Using reconstructed geometry and camera exterior orientations we can quickly model the scene (\textbf{c}). Finally, we pass all the parameters into a Differentiable Renderer (DR) for scene parameter learning (\textbf{d}). Note, camera alignment is only used for validation and not during the training or online phase.}

To enable real world experimentation, we recreate physical versions of our synthetic world setups. To enable object learning requires 3 steps, a) data acquisition, b) cameras alignment, c) object and scene modeling (\refFig{RealSetup}).

\textbf{Data acquisition} is performed using two standard Canon DSLR cameras fixed on tripods. We set the validation (novel-view) camera at an $\sim$90\textdegree
angle from the training camera view. To create the validation dataset objects are then placed at random within the scene and the scene is captured from each camera. A training dataset is then created by randomly placing objects into the scene and only capturing data from the training camera. We finally take an image from each camera with no objects in the scene for scene background modeling.

\textbf{Camera alignment}
To get relative exterior orientations between the two cameras we utilize a Structure-from-Motion (SfM) pipeline. Each camera is lifted from the tripod and multiple images take from various viewpoints. We ensure we densely cover the space between the two cameras. The images are input into the SfM software which produces the exterior orientations of the cameras in a common (arbitrary) coordinate system. We further generate a dense point cloud using multi-view stereo to aid object modeling. All SfM processing was performed in MetaShapes v1.6.5 using default settings.

\textbf{Object and scene modeling}
Each shape is modeled inside Blender software, based on the dense point cloud. This produces mesh objects which we pass to the DR. By doing so we ensure our mesh has the same scale as the scene.

We assume a flat ground plane and that objects are in a canonical position, with the only rotation axis being about the objects up-axis. We further transform the camera exterior orientations such that the reconstructed ground plane lies on the $XY$ plane. This is done manually as we have no ground control points to align the bundle adjustment. If a real co-ordinate system is required, ground control points could be placed into the scene avoiding the need for manual transformation (rotation and scaling) of the exterior orientations.

Once the camera alignment is complete, we can then create a ground plane mesh which we texture with the image containing no objects for training and validation cameras respectively. To ensure the same view of the ground plane we model the ground plane to the view frustum of the camera. This can be done by importing the intrinsic matrix estimated from the SfM to a virtual camera. Camera exterior and intrinsic properties are exported along with all mesh objects.

In scenarios where 3D CAD objects are available prior to data capture (e.g. industrial applications where parts are designed in CAD software before being manufactured), the object modeling process is not required.

\textbf{Bunny}
The \textsc{Bunny} dataset demonstrates a more casual data capture method. A single 3D-printed Stanford bunny is filmed from a moving mobile camera. Training data is generated by extracting frames every $n$-frames where $n=10$ in our case. Camera alignment is then computed for a sequence of frames. The first frame in the sequence is used as the input frame and the last frame is used as the novel view. As the motion is continuous, we can get relative camera positions using SfM. Object is then the same as for the \textsc{Real Sphere} and \textsc{Shapes} datasets. In order to model the scene, we require a background image, which contains no object. We found Adobe Photoshop content-aware delete tool to quickly and effectively remove the foreground object, leaving a clean background image which can be used to texture the ground-plane mesh.

\subsection{Validation Experiment}\label{sec:ValidationExperiment}
In this section we propose an experiment to verify the effect the critic has on the network optimization. Our key aim to construct an environment where the input data distribution is from a known distribution $\mathcal{D}$, and therefore the behavior of the critic (to ensure all network outputs belong to this distribution) can be evaluated. We therefore seek to find a differentiable analytical solution which calculates the distance between distribution of the current network output and $\mathcal{D}$. Such a solution would act as an Oracle critic $\mathcal{O}$ (i.e., a perfectly behaved critic).

\textbf{Setup} To aid the experiment, we first reduce our problem space to 2D, where our network must predict a position $t \in [0, 1]$ along the $x$-plane and a luminance $l \in [0, 1]$ (see \refFig{ParamSpace}).

\textbf{Data Distribution} To generate our training dataset, we sample parameters $p\in \mathbb{R}^{t, l}$ from a uniform distribution $\mathcal{D}: \{t, l\} \sim \mathcal{U}(0,1)$. Our optimization then becomes $$\argmin{} D(\hat{p} - \mathcal{D}), $$ where $D$ is a distance metric defined below.

\textbf{Oracle Critic} We define $D$ such that it allows for gradient-based parameter optimization. First, we note that a uniform distribution $\mathcal{U}$ would have a constant density across the parameter space. Therefore, we aim to penalize $\hat{p}$ varying from a constant density. This variation we refer to a discrepancy $D$. 

We can measure $D$ by testing for a low discrepancy pattern between all possible subsets of $\hat{p}$ and a random sampling $S \sim \mathcal{D}$. The size $N$ of sampling $S$ here is not strictly important, so long as it represents $\mathcal{D}$ sufficiently. In our experiments $N=300$.

First we calculate the density with a Kernel Density Estimate (KDE) $E$. Whilst it would be tempting to compute $E$ using common methods such as a box or Epechnikov approach, these would not be differentiable. Instead, we use a steep Gaussian kernel $$K = \exp(-(||\hat{p}-S||)^2).$$ Next, we compute KDE estimates $E$ for each point $e \in K$ as $$\frac{1}{2}(e_t^i + e_l^i).$$ Finally, we calculate the variance in the KDE response (discrepancy) $E$ as $$D = \frac{1}{N-1} \sum_{i=0}^{N}{(E_i - \bar{E})^2}.$$ This is the scalar value we minimize in the optimization.

When a simple L2 loss is used and no distribution is enforced (Fig. \ref{fig:ValExpIterations} top) the optimization almost always falls into the local-minima of turning the circle black ($l=0$). This is the same phenomenon we experience in the main paper for $\mathtt{NonCur}$ experiments. We observe this because the majority of gradients for $l$ point towards the black background. As a result, the overall gradient is dominated by this incorrect force. This is visible in the main paper Fig. 6 (left) as parameters falling towards the bottom of the chart. By enforcing $\hat{p} \sim \mathcal{D}$, we see that when the parameters start to fall, they are push back up, where they find more dominant L2 gradients main paper Fig. 6 (right).

A range of optimization visualizations can be found in the supplemental folder under $\mathtt{./val\_experiment}$.

\myfigure{ParamSpace}{Visualization of the parameter space when rendered.}

\subsection{Visualizing training}\label{sec:VisTraining}

In this section we visualize various training scenarios to show the effect of a real critic during training on simple problem. We present two scenarios where the parameter space is 5D ($x,y,r,g,b$) and 6D ($x,y,z,r,g,b$) respectively. For each scenario we demonstrate learning using image loss ($5D_{\mathtt{IMG}}$ / $6D_{\mathtt{IMG}}$), b) curiosity ($5D_{\mathtt{CUR}}$ / $6D_{\mathtt{CUR}}$) and image loss and curiosity ($5D_{\mathtt{OURS}}$ / $6D_{\mathtt{OURS}}$). We generate a dataset by randomly sampling position from a uniform distribution within a fixed bound and each color from a pre-computed color map. This allows us to observe if the critic is able to learn to enforce both the position and color distribution. By visualizing the step-by-step optimization, we note the following observations.

In the 5D scenario, $5D_{\mathtt{IMG}}$ follows dominant and incorrect gradients to match the background color. Furthermore, the gradients push the object towards the edge of the scene, and ultimately out of view of the camera which results in perfectly matching the background. $5D_{\mathtt{CUR}}$ learns to make predictions within the position and color distributions. As no image loss is present, there is no incentive for the network to make a prediction on a given single sample, but instead generate proposals that match the entire distribution. $5D_{\mathtt{OUR}}$ combines both the image loss (i.e., loss on a single given sample) and the critic loss (i.e., loss on the entire distribution). This results in avoiding the local and incorrect minima and results in convergence with respect to the position and color parameters. 

Similar observations are made in the 6D scenario. With the additional dimension $6D_{\mathtt{IMG}}$ also finds a local minima by reducing the size of the sphere. This is achieved in practice by pushing the sphere further away from the camera (i.e., $> z$). $6D_{\mathtt{CUR}}$ learns the new position distribution and generates plausible samples that comply with the dataset generation constraints. $6D_{\mathtt{OURS}}$ again finds the correct parameter solution for both position and color. Full training sequences can be found in the supplementary files under $\mathtt{./training\_vis/\{5d, 6d\}}$.

\myfigure{VisTraining}{Example outputs of training visualization experiments for 5D scenario. Columns 1-7 show progress during training. Columns 8 and 9 show the final models network output (col. 9) given an input (col. 8). Full videos can be found at: $\mathtt{./training\_vis/\{5d, 6d\}}$.}

\mycfigure{ValExpIterations}{Renderings of parameters at regular intervals during optimization for \textbf{a)} without curiosity and \textbf{b)} with curiosity. When image loss is used without a curiosity term the parameters degrade into an out-of-distribution solution where luminance $l$ tends towards 0. Optimization visualizations can be found in the supplementary files under $\mathtt{./val\_experiment}$.}

\mycfigure{Results2}{Additional qualitative results.}

\mycfigure{Results3}{Additional qualitative results.}

\mycfigure{Linemod1}{Additional qualitative results on LineMod dataset.}
\mycfigure{Linemod2}{Additional qualitative results on LineMod dataset.}

% {\small
% \bibliographystyle{icml2021}
% \bibliography{egbib}
% }

\end{document}